\documentclass[10pt,twocolumn,letterpaper]{article}

\usepackage{cvpr}
\usepackage{times}
\usepackage{epsfig}
\usepackage{graphicx}
\usepackage{amsmath}
\usepackage{amssymb}
\usepackage{subfig}
\usepackage[export]{adjustbox}
\usepackage{multirow}

\usepackage[pagebackref=true,breaklinks=true,letterpaper=true,colorlinks,bookmarks=false]{hyperref}

\cvprfinalcopy % *** Uncomment this line for the final submission

\def\cvprPaperID{20} % *** Enter the CVPR Paper ID here

\ifcvprfinal\fi
\begin{document}

%%%%%%%%% TITLE
\title{Symbol Spotting on Digital Architectural Floor Plans Using a Deep Learning-based Framework}

\author{Alireza Rezvanifar, Melissa Cote, Alexandra Branzan Albu\\
University of Victoria,\\
British Columbia, Canada\\
{\tt\small \{arezvani, mcote, aalbu\}@uvic.ca}
}

\maketitle
%\thispagestyle{empty}

%%%%%%%%% ABSTRACT
\begin{abstract}
This papers focuses on symbol spotting on real-world digital architectural floor plans with a deep learning (DL)-based framework. Traditional on-the-fly symbol spotting methods are unable to address the semantic challenge of graphical notation variability, i.e.\@ low intra-class symbol similarity, an issue that is particularly important in architectural floor plan analysis. The presence of occlusion and clutter, characteristic of real-world plans, along with a varying graphical symbol complexity from almost trivial to highly complex, also pose challenges to existing spotting methods. In this paper, we address all of the above issues by leveraging recent advances in DL and adapting an object detection framework based on the You-Only-Look-Once (YOLO) architecture. We propose a training strategy based on tiles, avoiding many issues particular to DL-based object detection networks related to the relative small size of symbols compared to entire floor plans, aspect ratios, and data augmentation. Experiments on real-world floor plans demonstrate that our method successfully detects architectural symbols with low intra-class similarity and of variable graphical complexity, even in the presence of heavy occlusion and clutter. Additional experiments on the public SESYD dataset confirm that our proposed approach can deal with various degradation and noise levels and outperforms other symbol spotting methods.
\end{abstract}

\section{Introduction}
\label{Sec:Intro}

Symbol spotting  \cite{rezvanifar2019symbol, rusinol2010symbol, santosh2018document} refers to the retrieval of graphical symbols embedded in larger images, typically in the form of a ranked list of regions of interest more likely to contain the symbols. Unlike symbol recognition, which aims to automatically label an already isolated symbol, spotting happens in context. It is typically carried out on the fly; no prior information about the shape of the symbols is known, and therefore machine learning-based methods are not helpful. This limitation can be in fact construed as a positive, as it eliminates the need for a training set. Annotated real-world datasets can be very difficult to obtain and few are publicly available; this is especially true for architectural floor plans, due to the intellectual property often restricting their use and publication, and to their sheer complexity and density of embedded information, which makes the annotation process a daunting task. On-the-fly symbol spotting circumvents the training process from annotated real-world datasets via an interactive process: the user crops a query patch in the image and the system finds all similar patches within the image based on the statistical and geometrical information of the query patch. It is assumed that the user-identified patch contains a symbol. 

\begin{figure}[t!]
\centering
\includegraphics[width=0.85\linewidth]{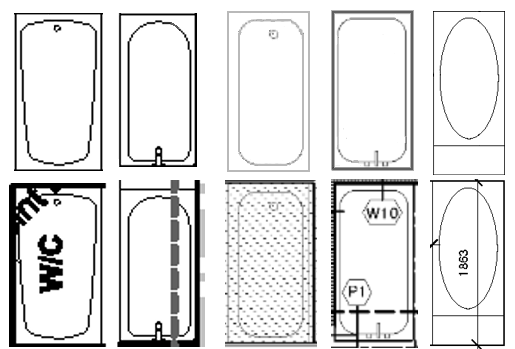}
\caption{First row: 5 different graphical notations of the bathtub symbol. Second row: corresponding symbol instances in real-world scenarios with occlusions, clutter and various levels of degradation.}
\label{Fig:NotationVar}   
\end{figure}

One crucial drawback of on-the-fly symbol spotting is that it cannot cope with graphical notation variability. Being able to deal with such variability is very important in the context of designing a scalable method which is applicable to various semantically equivalent graphical representations. This is particularly true for architectural floor plans, as there can be as many graphical notations for a given symbol as there are architectural firms, and even more. Fig.\@ \ref{Fig:NotationVar} (top row) illustrates some of the graphical notation variability for the bathtub symbol. In this paper, we relax the ``on-the-fly" property of traditional symbol spotting and instead tackle this semantic challenge by proposing a deep learning-based method that is scalable to various semantically equivalent graphical representations.

Another important consideration is the presence of various levels of occlusion and clutter in architectural plans. Architectural floor plans, as scale-accurate two-dimensional diagrams of one level of a building, consist of lines, symbols, and textual markings, showing the relationships between rooms and all physical features with an elevated view from above. In real-world plans, the quantity of information that has to be conveyed by architects for the proper construction or renovation of the building is significant, yielding often to heavy occlusion and clutter. Fig.\@ \ref{Fig:NotationVar} (bottom row) shows instances of bathtub symbols as they appear within architectural floor plans, suffering from heavy clutter and occlusion. Such occlusion, clutter and degradation can strongly degrade the performance of symbol spotting systems. If, as a result, the shape of the symbols appears considerably distorted, state-of-the-art symbol spotting methods cannot detect the degraded symbols. In this paper, we aim to provide a method that is robust to heavy occlusion and clutter.

\begin{figure}[t!]
\centering
\includegraphics[scale=0.3]{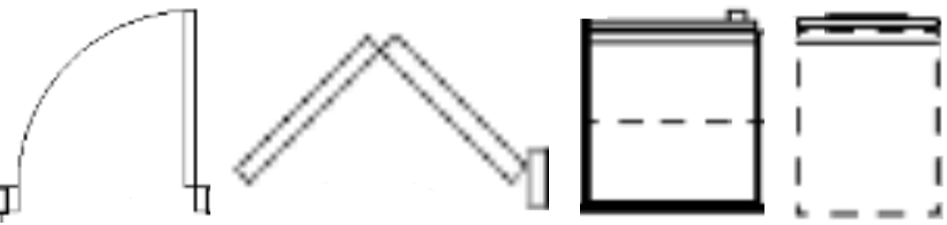}
\caption{Some examples of trivial symbols consisting of few and less informative primitives (from left: entry door, closet door, refrigerator and dishwasher).}
\label{Fig:SimpleSyms}   
\end{figure}

A third issue relates to the graphical simplicity of symbols. Simple (trivial) symbols that do not have complex structures, such as those shown in Fig.\@ \ref{Fig:SimpleSyms}, can be challenging for many traditional symbol spotting methods. As can be seen from the figure, the constituent primitives of these symbols are limited and structural-based methods cannot extract well-informed descriptions. In this paper, we successfully address the detection of symbols of varying graphical complexity (from very simple to highly complex).

\subsection{Contributions}
\label{Sec:Intro_Cont}
This paper proposes a DL-based framework for spotting symbols in digital real-world architectural floor plans. Our contributions are two-fold. 
\begin{enumerate}
\item We leverage recent advances in DL by adapting a YOLO-based \cite{redmon2017yolo9000} object detection network to the problem of symbol spotting in real-world architectural floor plans. We propose a training strategy based on tiles, which allows us to circumvent many issues particular to DL object detection networks, including the size of the objects of interest relative to the size of the images, aspect ratios, and data augmentation.
\item Our proposed DL-based symbol spotting framework successfully addresses the main issues of traditional on-the-fly symbol spotting, namely graphical notation variability, occlusion and clutter, and variable graphical complexity of symbols.
\end{enumerate}

The remainder of the paper is structured as follows: Sec.\@ \ref{Sec:LitReview} reviews related works, Sec.\@ \ref{Sec:ProposedMethod} details our symbol spotting approach, Sec.\@ \ref{Sec:Results} discusses experimental results, and Sec.\@ \ref{Sec:Conclusion} presents concluding remarks.

\begin{figure*}[t!]
\centering
\includegraphics[width = 0.92\textwidth]{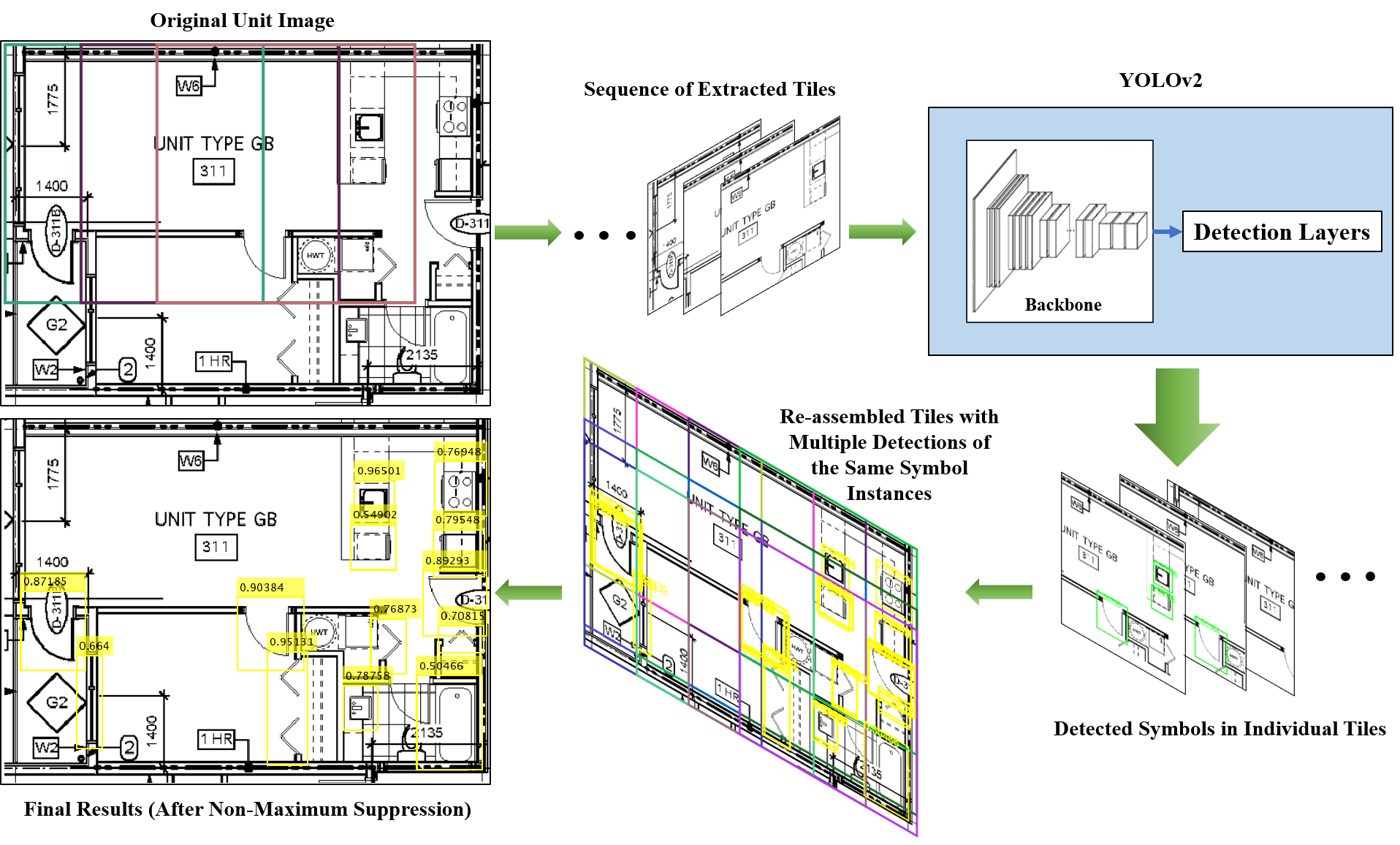} 
\caption{Proposed symbol spotting framework. Overlapping tiles from the input image are passed through YOLOv2 and individually processed. Non-maximum suppression is carried out to remove multiple detections of the same symbol instances due to the tiling strategy.}
\label{Fig:TileDetector}
\end{figure*}

\section{Related Works}
\label{Sec:LitReview}
 
Traditional (i.e.\@ non-DL) symbol spotting approaches can be categorized as either pixel-based or vector-based, depending on the type of primitives used for representing symbols. They both typically involve two phases: a description phase, in which low level information is utilized to construct a symbol descriptor, and a matching phase, in which regions within the image that match the symbol description are found \cite{rezvanifar2019symbol}. Pixel-based approaches work directly on the raster image format, and are usually associated with statistical methods, while vector-based approaches require a conversion to vectorial primitives and are usually associated with structural methods (typically graph-related).

Examples of pixel-based approaches include the F-signature \cite{tabbone2003matching}, which describes symbols based on exerted attraction forces between different segments; methods based on pixel-level geometric constraints (e.g.\@ \cite{yang2005symbol}) summarized in histograms and matched via histogram similarity; the Blurred Shape Model (BSM) \cite{escalera2009blurred}, which encodes the spatial probability of occurrence of shapes from the skeleton image and a neighbourhood-based voting system; its extension (Circular Blurred Shape Model, CBSM) \cite{escalera2011circular}, which utilizes correlograms to make the description rotational invariant; and the Shape Context for Interest
Points (SCIP) \cite{nguyen2008symbol} as well as its extension ESCIP \cite{nguyen2009symbol}, describing symbols with visual words. One important drawback of pixel-based methods is their high computational complexity, which results in a slow matching phase.

Vector-based approaches start by constructing a vectorial representation of the meaningful parts of images and symbols using grouped constituent primitives. Examples of constituent primitives include vectors and quadrilaterals \cite{ramel2000structural}; solid components, circles, corners, and extremities \cite{santosh2012symbol, santosh2014integrating, santosh2014bor}; critical points and lines \cite{broelemann2012hierarchical,broelemann2013hierarchical}; convex regions \cite{dutta2013near}; contour maps \cite{nayef2011statistical}, closed regions \cite{rusinol2007region, le2009symbol, le2012integer}; and image regions \cite{barducci2012object}, derived for instance from connected components. Spatial relationships between primitives are then typically encoded in structural graphs. Examples of such graphs are the Full Visibility Graph (FVG) \cite{locteau2007symbol}, which focuses on the existence of a straight line between two points on the primitives such that the line does not touch any other primitive; the Attributed Relational Graph (ARG), which qualifies the type of intersection \cite{ramel2000structural, qureshi2007spotting} or connections \cite{santosh2012symbol, santosh2014integrating} between the primitives; the Hierarchical Plausibility Graph (HPG) \cite{broelemann2012hierarchical,broelemann2013hierarchical}, which tackles different possible vectorization errors; and the Region Adjacency Graph (RAG) \cite{le2009symbol,le2012integer,barducci2012object,dutta2013near}, characterizing region boundaries and the relational information between regions. In the matching phase, subgraph isomorphism is generally carried out to determine whether the image graph contains a subgraph isomorphic to the symbol graph. As graph matching techniques are computationally expensive, alternative matching methods have been proposed, such as graph serialization \cite{dutta2013symbol}, graph embedding \cite{luqman2011subgraph}, and relational indexing \cite{rusinol2010relational2}. One drawback of vector-based methods is the need for an initial vectorization, which can introduce errors and inaccuracies in the representation. The spatial relationships between primitives are also typically limited to very specific information (e.g.\@ adjacency in RAG, visibility in FVG).

Coping with notation variability of symbols remains a significant semantic challenge for traditional symbol spotting methods. Indeed, although some methods are relatively successful in dealing with noise, occlusion and clutter in the image \cite{broelemann2013hierarchical, rusinol2010relational2, santosh2014integrating}, they are not capable of detecting symbols with low intra-class similarity. Non-traditional, DL-based methods have only recently started to permeate the literature and are still far from addressing the current issues of traditional symbol spotting approaches, as they  mostly target only symbol recognition applications. For instance, in \cite{dey2017shallow}, the authors propose a shallow CNN for recognizing hand-drawn symbols in the context of multi-writer scenarios. In \cite{riba2017graph}, the authors utilize a message passing neural network (MPNN), which is a graph neural network, to globally describe the structural graph representation of symbols and then use the output for graph matching. Testing is limited to symbol recognition, as localization in context is problematic. Also, MPNNs are typically useful for dense graphs and do not yield the same performance on sparse graphs, which are common for our application domain. More recently, Ghosh et al.\@ \cite{ghosh2019gsd} proposed GSD-Net, a compact network for pixel-level graphical symbol detection. They use a semantic segmentation network, which labels all pixels individually as opposed to extracting bounding boxes around objects of interest. Such a method requires expensive pixel-level annotations. The authors also trained their network on the public SESYD dataset \cite{valveny2013report}, which is much simpler than real-world architectural floor plans. In particular, SESYD does not include occlusion, clutter, nor any symbol intra-class graphical variability. Closer to our work, Ziran and Marinai \cite{ziran2018object} and Goyal et al.\@ \cite{goyal2019bridge} both utilized object detection networks for symbol spotting. Their experiments, focused on floor plans significantly simpler than ours (see Sec.\@ \ref{Sec:Results_Real}), did not allow for a performance assessment under heavy occlusion and clutter such as that shown in Fig.\@ \ref{Fig:NotationVar}.

\section{Proposed Method}
\label{Sec:ProposedMethod}

The recent success of DL-based systems and convolutional neural networks (CNNs) has revolutionized the object detection field. Popular networks such as Single Shot Multibox Detector (SSD) \cite{liu2016ssd}, You Only Look Once (YOLO) \cite{redmon2016you, redmon2017yolo9000, redmon2018yolov3} and Faster R-CNN \cite{ren2015faster} can be used to detect thousands of classes in natural scenes. Their success is due in large part to the existence of large annotated datasets such as Pascal VOC \cite{everingham2010pascal}, MS COCO \cite{lin2014microsoft}, and ImageNet \cite{deng2009imagenet}.

In this work, we first build a dataset of real-world architectural floor plans. We then use this dataset to train a single shot detector based on the YOLOv2 \cite{redmon2017yolo9000} architecture for spotting architectural symbols within architectural floor plan images. Fig.\@ \ref{Fig:TileDetector} offers an overview of our proposed framework. The dataset preparation and our approach based on YOLOv2 are presented in detail next.

\subsection{Dataset Preparation}
\label{Sec:ProposedMethod_Prep}

\begin{figure}[t!]
\centering
\includegraphics[scale=0.60]{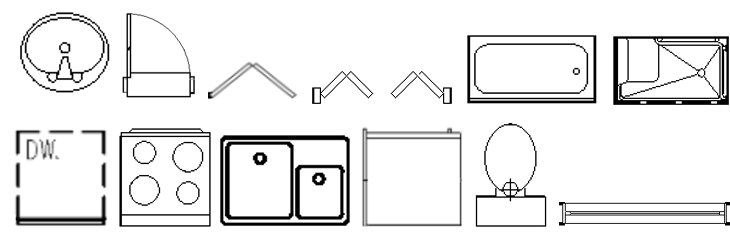}
\caption{Examples of each symbol class. First row from left: bathroom sink, entry door, single folding door,  double folding door, bathtub, shower. Second row: dishwasher, range, kitchen sink, refrigerator, toilet, and window.}
\label{Fig:QSymbols}   
\end{figure}

From a library of proprietary digital architectural drawings, designed by 10 architectural firms, we selected 115 different units showing various levels of difficulty in terms of density of visual information, occlusion, and clutter. Architects typically share floor plans in the PDF format. We converted the PDFs into 150 DPI images, and annotated 12 architectural symbol classes, such as bathroom sinks, windows, and entry doors (see Fig.\@ \ref{Fig:QSymbols}). We do not make the dataset of real-world architectural plans public due to intellectual property issues, but are working towards securing the necessary permissions for a future release.

\begin{figure}[t!]
\centering
\includegraphics[scale=0.42]{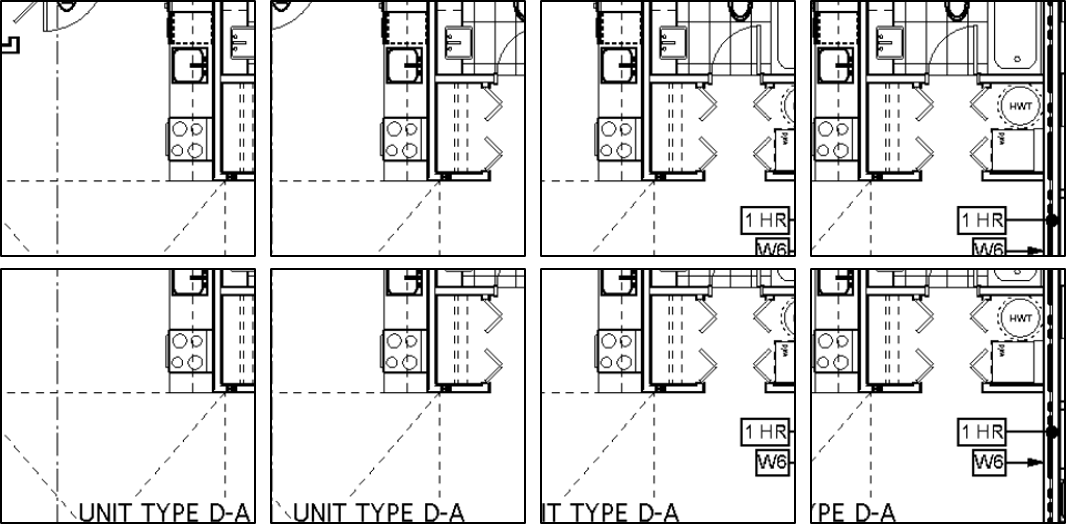}
\caption{Data augmentation via image tiling strategy. The range symbol appears at various locations within the tiles, which also include various other symbols.}
\label{Fig:DataAug}   
\end{figure} 

We face several problems when dealing with architectural floor plan images in the context of DL systems. First, the average floor plan size is $5400 \times 3600$ pixels, whereas individual symbols are very small (e.g.\@ $70 \times 80$ pixels for a bathroom sink). As a result, symbols tend to disappear in the output feature map of CNNs, preventing them from being detected. In addition, floor plan images have diverse aspect ratios and resizing them to a fixed size, as required by CNN architectures, dramatically changes the symbol morphology and thus decreases the classification performance.

We propose a tiling strategy to tackle the above problems, which uses a scale parameter $\alpha$ and stride size $S$. First, all the $[\alpha M\times \alpha M]$ overlapping tiles that have a starting point at least $S$ pixels apart are extracted from the floor plan images. $[M\times M]$ is the required input size of the utilized CNN, which is usually less than $[256 \times 256]$ \cite{redmon2016you}. Tiles that do not encompass at least one complete symbol are automatically discarded from the training dataset. The tile size must be selected so that tiles are larger than symbols. Also, selecting larger tiles can boost their contextual information, as in architectural plans, the occurrences of some symbols might be spatially correlated. For instance, we can expect to see a bathroom sink symbol in the vicinity of a toilet symbol. At the same time, the tiles must be small enough so that the symbols still appear in the deeper layers and output of the CNNs. Tiles are also useful for data augmentation. Fig.\@ \ref{Fig:DataAug} shows neighbouring tiles containing a range symbol captured at different locations within the tiles. The tile size in the figure is $[224 \times 224]$ (i.e. \@ $\alpha = 1$ and $M=224$, required by ResNet50 \cite{he2016deep}) and $S = 50$ pixels.

\subsection{Symbol Spotting Using YOLOv2}
\label{Sec:ProposedMethod_Training}

Single shot object detection architectures based on image grids (such as YOLO) seem appropriate and accurate enough to localize architectural symbol boundaries, compared to more complex and heavier two-stage classification architectures (such as Faster R-CNN \cite{ren2015faster}), due to the following considerations. Floor plan images differ from natural scene images (for which most CNN-based object detection systems were developed) on several aspects. Floor plans are typically grey-level with a small number of possible symbol classes, compared to colourful natural scene images with a large number of possible object classes. Additionally, floor plans have a simpler background and chances of overlap between symbols is low (this does not apply of course to other parts of the image such as textual information and measurements, which may have a considerable overlap with symbols). Here, we use YOLOv2 \cite{redmon2017yolo9000} as a single shot object detection architecture. We selected YOLOv2 instead of YOLO \cite{redmon2016you} because of its higher localization accuracy and recall performance. The improvements of YOLOv3 \cite{redmon2018yolov3} consist mainly in a prediction across three scales and a better feature extraction network, but at the cost of a slower and heavier system. As architectural symbols have similar sizes and simpler structures compared to objects in natural scenes, YOLOv3 cannot offer a noticeable improvement over YOLOv2. In YOLOv2, the input image is divided into non-overlapping grids. Each grid can only predict one object. A backbone CNN (e.g.\@ Darknet19) extracts features and for each grid, a set of prior anchors are evaluated based on a loss function which penalizes localization, confidence and classification errors.

In the training phase, we use the tile dataset described in Sec.\@ \ref{Sec:ProposedMethod_Prep}) to train the network. In the inference phase, the input image is first broken down into tiles. Each tile is then passed through the network and symbols are detected. Fig.\@ \ref{Fig:TileDetector} shows the inference process. The detected symbols in the overlapping tiles are shown in the bottom-right image. As a given instance of a symbol typically appears in several tiles, it is detected multiple times. To refine and concatenate the results, we perform a non-maximum suppression step as follows. For overlapping detections, we compare all pairs of bounding boxes. If their overlap is larger than a threshold (a percentage of the size of the smaller bounding box), the bounding box with the highest classification score is retained. In cases of close scores, the larger bounding box is selected and the smaller one is removed. The bottom-left image in Fig.\@ \ref{Fig:TileDetector} shows the final results for a $10\%$ threshold.

\section{Results and Discussion}
\label{Sec:Results}

We assess our framework on a real-world floor plan dataset and on SESYD, a public dataset of synthetic documents. For both datasets, we evaluate the performance on individual tiles first, and then assess entire floor plans. We provide a comparative analysis of our approach with respect to state-of-the-art symbol spotting methods for SESYD only, as code implementations of these methods are either unavailable or not functional on our real world dataset.

\subsection{Real-World Images}
\label{Sec:Results_Real}

From the 115 units of the dataset (see Sec.\@ \ref{Sec:ProposedMethod_Prep}), we used 90 units for extracting tiles. The remaining 25 units are used as a test set for evaluating the framework on entire floor plans. Given $S = 50$ and a tile size of $[227\times227]$, the 90 units generated 4707 tiles containing at least one complete symbol. We randomly selected 80\% of those 4707 tiles for training the network, with the remaining 20\% tiles used for validation. During training, we employed the Adam optimizer \cite{kingma2014adam} with a mini-batch size of 30, a fixed learning rate of $10^{-4}$, and data augmentation with horizontal and vertical flipping and rotation and scale changes. Moreover, 10 prior anchors were calculated from the size of the symbols. We experimented with three different backbones, the original Darknet19 \cite{redmon2017yolo9000}, as well as ResNet50 \cite{he2016deep} and Xception \cite{chollet2017xception}.

\begin{table}[h!]
\centering
\caption{Performance evaluation on the tile validation set for two datasets and different backbones.}
\label{tab:backboneEva}
\begin{tabular}{l|l|l|l|l}
\noalign{\smallskip}
\textbf{Dataset} & Backbone & $mAP$ & $AP_{50}$ & $AP_{75}$ \\
\noalign{\smallskip}\hline\hline\noalign{\smallskip}
\multirow{3}{*}{Real-world} & ResNet50 \cite{he2016deep} & 72.40 & 96.20 & 90.13 \\
 & Darknet19 & 61.53 & 93.7 & 72.41 \\
 & Xception \cite{chollet2017xception} & 51.03 & 87.58 & 55.01 \\
\hline
SESYD & ResNet50 & 78.15 & 97.92 & 91.42 \\
\hline
\noalign{\smallskip}
\end{tabular}
\end{table}

\begin{table}[h!]
\centering
\caption{Performance evaluation per symbol class and globally on the real-world test dataset for different backbones.}
\label{tab:backboneEvaRealWorld}
\begin{tabular}{l|l|l|l|l}
\noalign{\smallskip}
\textbf{Symbol} & \multicolumn{2}{l|}{ResNet50} & \multicolumn{2}{l}{Darknet19}\\
 \cline{2-5}
 & $AP_{50}$ & $AP_{75}$ & $AP_{50}$ & $AP_{75}$\\

\noalign{\smallskip}\hline\hline\noalign{\smallskip}
Bathtub & 91.67 & 91.67 & 95.83 & 95.83\\
\hline
Toilet & 100.00 & 50.87 & 100.00 & 77.27\\
\hline
Kitchen Sink & 91.07 & 77.91 & 88.21 & 51.74\\
\hline
Bathroom Sink & 84.97 & 56.63 & 83.02 & 46.73\\
\hline
Closet Door & 86.96 & 39.47 & 91.30 & 50.43\\
(double) & & & &\\
\hline
Entry Door & 86.35 & 82.47 & 89.81 & 83.60\\
\hline
Oven & 100.00 & 95.83 & 91.67 & 87.50\\
\hline
Window & 75.64 & 31.78 & 77.75 & 33.23\\
\hline
Refrigerator & 87.50 & 76.38 & 91.49 & 66.79\\
\hline
Closet Door & 88.76 & 59.99 & 95.46 & 23.77\\
(single) & & & &\\
\hline
Dishwasher & 78.89 & 66.35 & 67.00 & 67.00\\
\hline
Shower & 100.00 & 100.00 & 100.00 & 100.00\\
\hline
\textbf{AP} & \textbf{89.32} & \textbf{69.11} & \textbf{89.30} & \textbf{65.32}\\
\hline
\textbf{mAP} &  \multicolumn{2}{l|}{\textbf{59.03}} & \multicolumn{2}{l}{\textbf{56.50}}\\
\hline
\noalign{\smallskip}
\end{tabular}
\end{table}

\begin{figure}[t!]
\centering
\includegraphics[scale=0.25]{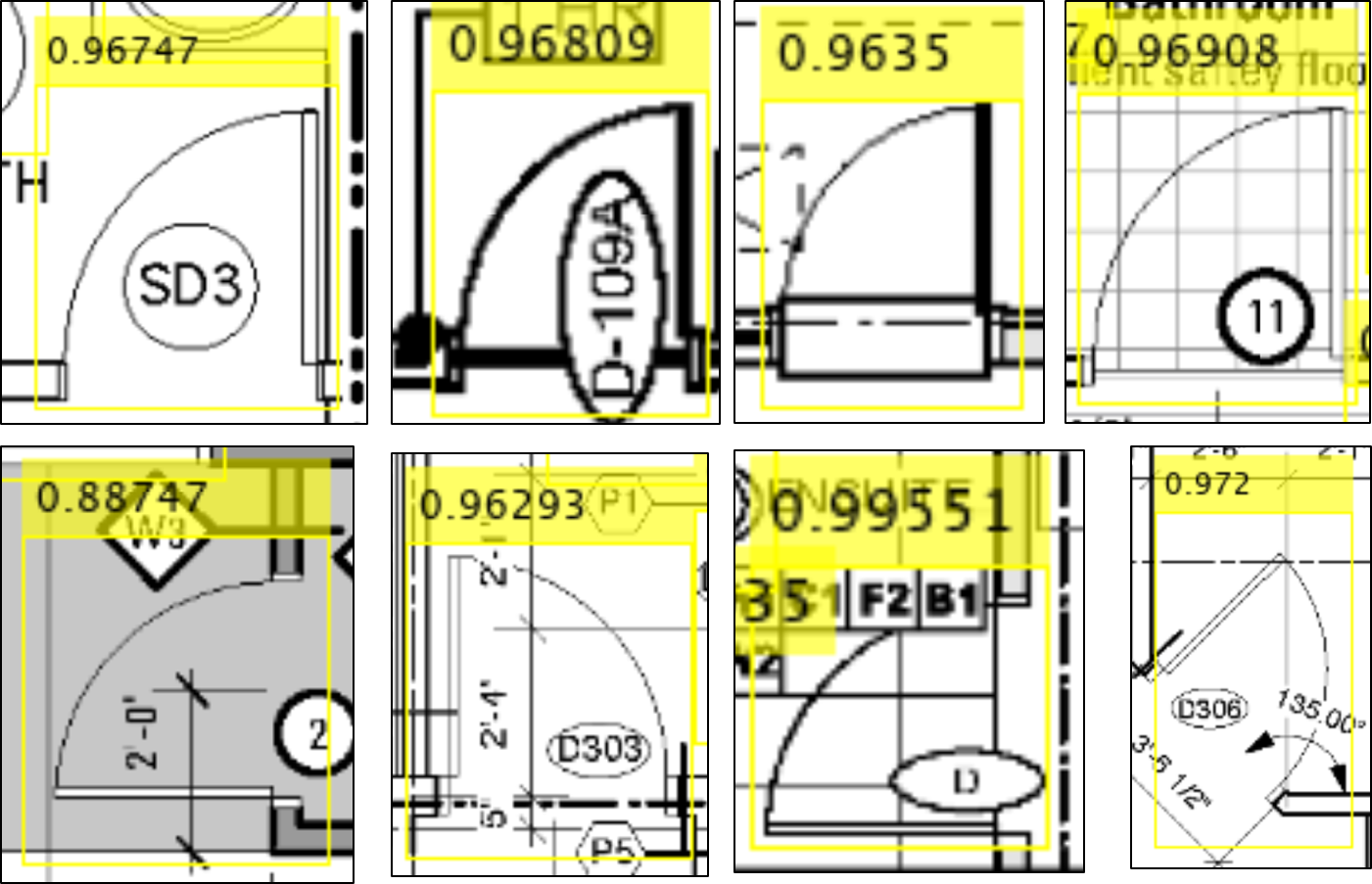}
\caption{Detected entry doors and scores (max = 1) for various levels of occlusion and overlap, in real-world dataset.}
\label{Fig:DetectedDoors}   
\end{figure} 

\begin{figure*}[h!]
\centering
\subfloat[]{\includegraphics[width = 0.46\textwidth, frame]{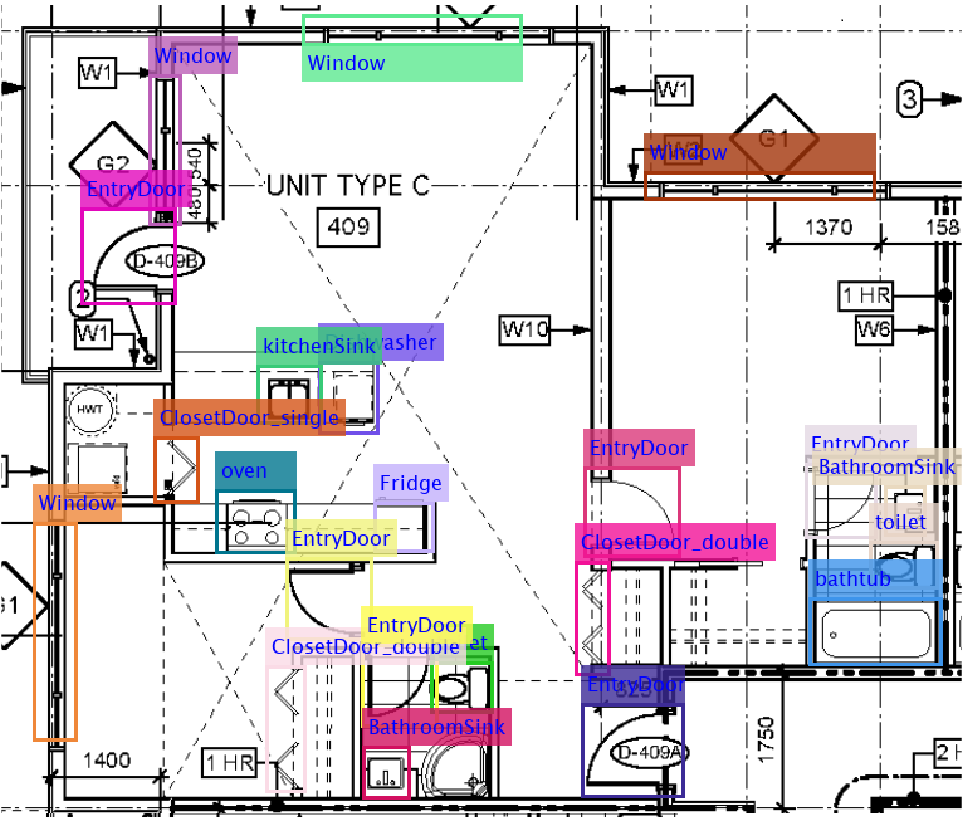}} 
\hspace{0.2em}
\subfloat[]{\includegraphics[width = 0.45\textwidth, frame]{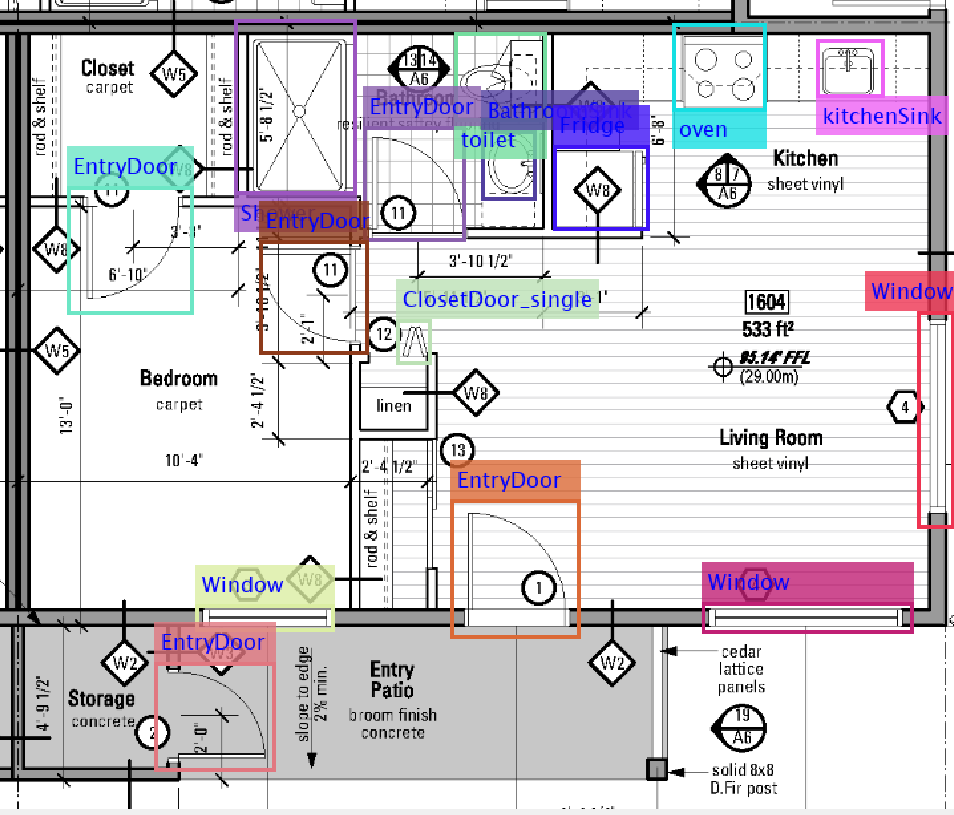}} \\
\subfloat[]{\includegraphics[width = 0.43\textwidth, frame]{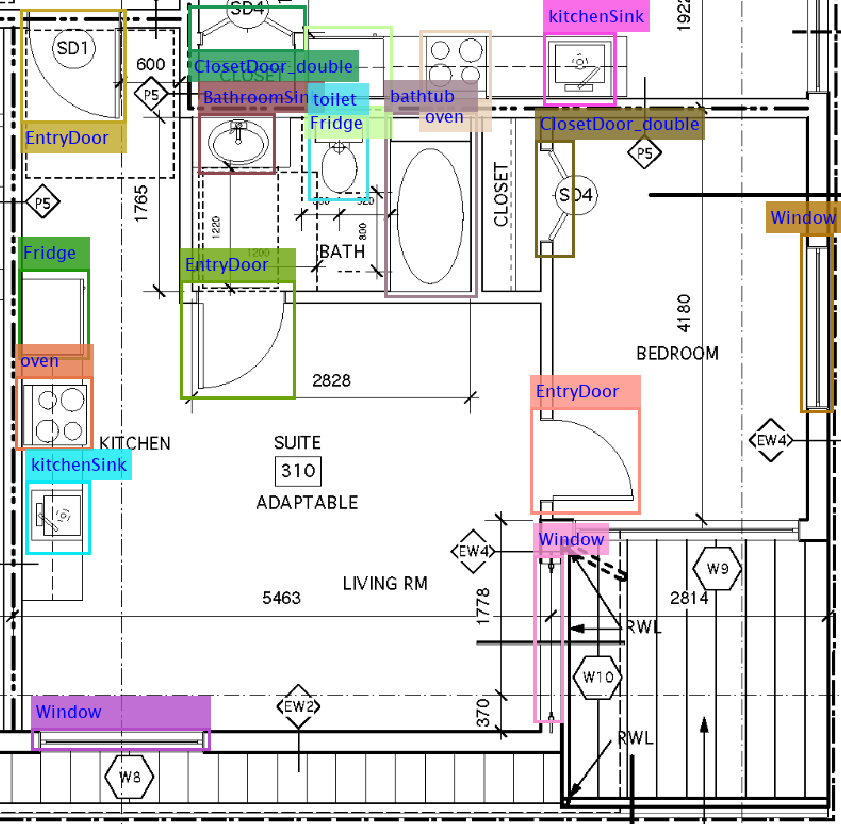} }
\hspace{0.2em}
\subfloat[]{\includegraphics[width = 0.46\textwidth, frame]{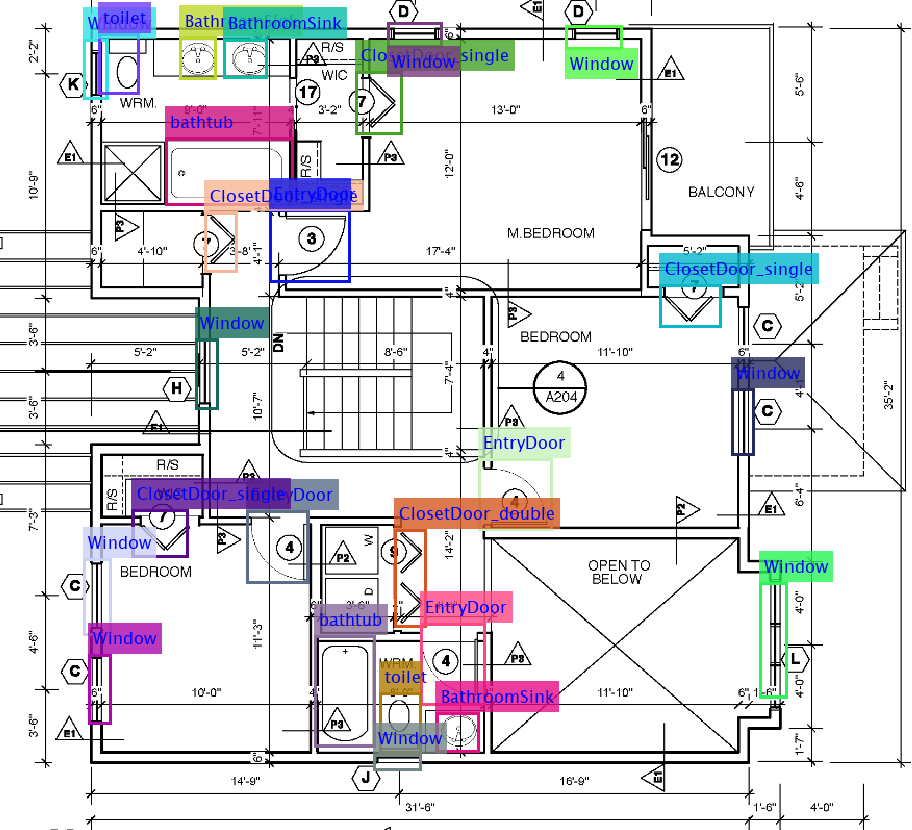} }\\
\caption{Examples of spotted symbols in real-world floor plan images.}
\label{Fig:FloorPlans}
\end{figure*}

Table \ref{tab:backboneEva} (first three rows) shows the performance on the tile validation set, whereas Table \ref{tab:backboneEvaRealWorld} shows the performance per symbol class and the global performance for the test set of 25 entire floor plans. In the tables, $mAP$, $AP_{50}$ and $AP_{75}$ represent the mean average precision and the average precision for IoUs equal to 50\% and 75\%, respectively. The IoU (Intersection over Union) is obtained as follows:

\begin{equation}
IoU(A,B) = | A \cap B | \: / \: | A \cup B |
\label{Eq:IoU}
\end{equation}
where $A$ and $B$ are the bounding boxes of the detected symbol and the ground truth symbol. From Table \ref{tab:backboneEva}, we can see that the ResNet50 backbone significantly outperforms Darknet19 and Xception, with Xception having the lowest performance. Looking specifically at the $AP_{50}$ metric, as an IoU of 50\% is acceptable in symbol spotting, the average precision is very high. From Table \ref{tab:backboneEvaRealWorld}, again focusing on $AP_{50}$, we can see that our method performs strongly for most symbols, with some yielding 100\% precision. The lowest score is obtained for the window symbol, which is a particularly difficult case due to its triviality and varying aspect ratio. Incorporating contextual information on walls could help improve the window detection results.

\begin{table*}[h!]
%\scriptsize
\centering
\caption{Performance evaluation per symbol class and globally on the GREC contest test dataset (from SESYD).}
\label{tab:SESYDeva}
\begin{tabular}{l|l|l|l|l|l|l|l|l}
\noalign{\smallskip}
\textbf{Symbol} & \multicolumn{2}{l|}{Ideal} & \multicolumn{2}{l|}{Noise 1} & \multicolumn{2}{l|}{Noise 2} & \multicolumn{2}{l}{Noise 3} \\
\cline{2-9}
 & $AP_{50}$ & $AP_{75}$ & $AP_{50}$ & $AP_{75}$ & $AP_{50}$ & $AP_{75}$ & $AP_{50}$ & $AP_{75}$\\
\noalign{\smallskip}\hline\hline\noalign{\smallskip}
armchair & 87.10 & 63.96 & 88.89 & 61.85 & 90.48 & 54.37 & 100.00 & 90.97\\
\hline
bed & 100.00 & 100.00 & 92.11 & 92.11 & 89.47 & 56.64 & 94.74 & 92.03\\
\hline
door1 & 100.00 & 34.73 & 100.00 & 48.34 & 100.00 & 53.68 & 100.00 & 53.62\\
\hline
door2 & 100.00 & 100.00 & 100.00 & 100.00 & 0.00 & 0.00 & 100.00 & 100.00\\
\hline
sink1 & 100.00 & 0.00 & 100.00 & 0.00 & 100.00 & 0.00 & 100.00 & 0.00 \\
\hline
sink2 & 98.86 & 98.86 & 98.38 & 98.38 & 100.00 & 62.39 & 100.00 & 100.00 \\
\hline
sink3 & 82.35 & 82.35 & 92.31 & 88.46 & 100.00 & 100.00 & 95.83 & 91.30\\
\hline
sink4 & 100.00 & 32.11 & 100.00 & 60.24 & 100.00 & 39.13 & 100.00 & 47.87\\
\hline
sofa1 & 100.00 & 76.91 & 100.00 & 34.97 & 97.30 & 57.44 & 98.08 & 54.48\\
\hline
sofa2 & 100.00 & 96.48 & 100.00 & 65.72 & 100.00 & 82.02 & 100.00 & 45.08\\
\hline
table1 & 100.00 & 20.62 & 100.00 & 22.03 & 100.00 & 15.02 & 100.00 & 15.04\\
\hline
table2 & 100.00 & 42.00 & 100.00 & 81.08 & 100.00 & 45.63 & 100.00 & 40.38\\
\hline
table3 & 100.00 & 72.02 & 100.00 & 38.34 & 100.00 & 100.00 & 100.00 & 100.00\\
\hline
tub & 100.00 & 100.00 & 95.00 & 78.62 & 100.00 & 71.33 & 100.00 & 74.69\\
\hline
window1 & 62.20 & 0.00 & 58.27 & 0.00 & 71.01 & 0.00 & 59.31 & 0.00\\
\hline
window2 & 11.65 & 0.00 & 36.78 & 0.00 & 35.93 & 0.00 & 13.21 & 0.00\\
\hline
\textbf{AP} & \textbf{90.13} & \textbf{57.50} & \textbf{91.36} & \textbf{54.38} & \textbf{89.30} & \textbf{65.32}& \textbf{91.32} & \textbf{56.59}\\
\hline
\textbf{mAP} & \multicolumn{2}{l|}{\textbf{54.08}} &  \multicolumn{2}{l|}{\textbf{53.93}} & \multicolumn{2}{l|}{\textbf{47.27}}& \multicolumn{2}{l}{\textbf{54.85}}\\
\hline
\noalign{\smallskip}
\end{tabular}
\end{table*}

Fig.\@ \ref{Fig:DetectedDoors} shows examples of detected entry doors, using the ResNet50 backbone, with the bounding boxes and detection scores (max = 1) highlighted. This figure showcases the efficiency of our DL-based symbol spotting system compared to the traditional methods. Our system successfully addressed occlusion and boundary degradation, which can highly affect the raster-to-vector conversion and thus the structural representation of symbols in methods such as \cite{dutta2013symbol, locteau2007symbol, santosh2012symbol}, rotation, which is one of the weaknesses of the pixel-based methods such as \cite{escalera2009blurred, weber2012symbol}, and graphical notation variability. Furthermore, as entry doors have a limited number of primitives, some of them cannot survive the vectorization step required by vector-based methods. They also include very small closed regions that can easily make the symbol unrecognizable by methods that employ closed regions as constituent primitives, such as \cite{le2012integer, rusinol2007region, rusinol2010relational2}. 

Fig.\@ \ref{Fig:FloorPlans} shows symbol spotting results for four units with different designs and layouts, using the ResNet50 backbone. Qualitatively speaking, the results are excellent, and we see that our approach works well even in the presence of high levels of noise, occlusion and image degradation. Considering the varied sources of the plans, we can also conclude that our method successfully bridges the semantic gap related to intra-class graphical notation variability.

\subsection{SESYD Dataset}
\label{Sec:Results_SESYD}

We also provide an evaluation on the public synthetic images of the Systems Evaluation SYnthetic Documents (SESYD)\footnote{http://mathieu.delalandre.free.fr/projects/sesyd/} dataset, which is the standard dataset in the field. Its synthetic floor plan collection includes 1000 floor plan images (some of which have very similar unit layouts), containing up to 16 different query symbol classes, with only one graphical notation per symbol class. For training the system, we randomly picked 50 floor plan images and extracted the tile images. Since images are large and the floor plans are sparser than real-world floor plans, we used $[680 \times 680]$ tiles with $S=100$ to include more contextual information around each symbol. This yielded 11,753 images divided into subsets of 9402 and 2351 tiles for training and validation purposes, respectively. To test our system on entire images, we used the selection from the GREC Symbol Recognition and Spotting contest \cite{valveny2011report}. This contest set contains 20 images from the original dataset of 1000 images (ideal) and three degraded versions (60 images). Noise levels \#1 to \#3 in \cite{valveny2011report} simulate thinner and thicker lines than the original image lines, and add global noise to the image, respectively. All of our results on SESYD are obtained with the ResNet50 backbone, as it yields a better performance on the real-world dataset.

\begin{table}[t!]
\scriptsize
\centering
\caption{Instance- and pixel-wise evaluation of symbol spotting approaches on SESYD.}
\label{tab:PerformanceEva}
\begin{tabular}{l|l|l|l|l|l}
%\hline\noalign{\smallskip}
Method & Eval.  & $P$ & $R$ & $F$ & Queries\\
\noalign{\smallskip}\hline\hline\noalign{\smallskip}
Nguyen \etal\@ ~\cite{nguyen2009symbol} & Instance & 70.00 & 88.00 & 79.50 &  6 \\
\hline
Broelemann \etal\@ ~\cite{broelemann2013hierarchical} & Instance & 75.17 & 93.17 & 83.21 &  All \\
\hline
Dutta \etal\@ \cite{dutta2013near} & Instance & 62.33 & 95.67 & 75.50 & All\\
\hline
Le Bodic \etal\@ ~\cite{le2012integer} & Instance & 90.00 & 81.00 & 85.30 & All\\
\hline
Nayef and Breuel ~\cite{nayef2011use} & Instance & 98.90 & 98.10 & 98.50 & 12\\
\hline
Winner in \cite{valveny2011report} (ideal) & Pixel & 62.00 & 99.00 & 76.00 & All \\
\hline
Winner in \cite{valveny2011report} (noise 1) & Pixel & 64.00 & 98.00 & 77.00 & All \\
\hline
Winner in \cite{valveny2011report} (noise 2) & Pixel & 62.00 & 93.00 & 74.00 & All \\
\hline
Winner in \cite{valveny2011report} (noise 3) & Pixel & 57.00 & 98.00 & 72.00 & All \\
\hline
Proposed method & Instance & 98.56 & 97.31 & 97.93 & \\
\cline{2-5}
(ideal) & Pixel & 77.35 & 98.97 & 86.83 & All\\
\hline
Proposed method & Instance & 99.32 & 97.15 & 98.22 & \\
\cline{2-5}
(noise 1) & Pixel & 77.69 & 97.28 & 86.39 & All\\
\hline
Proposed method & Instance & 99.11 & 99.11 & 99.11 & \\
\cline{2-5}
(noise 2) & Pixel & 76.48 & 97.65 & 85.78 & All\\
\hline
Proposed method & Instance & 77.63  & 97.34 & 96.62 & \\
\cline{2-5}
(noise 3) & Pixel & 99.46 & 93.93 & 96.60 & All\\
\noalign{\smallskip}\hline
\end{tabular}
\end{table}

Table \ref{tab:backboneEva} (last row) shows the results on the validation tile set, and Table \ref{tab:SESYDeva} shows the performance per symbol class and the global performance for the test set of 80 entire floor plans. Looking at $AP_{50}$ in both tables, our framework yields a very high precision rate, with 100.00\% for many of the symbol classes. Again, the window classes (window1 and window2) are the most problematic ones, and would benefit from additional contextual information.

Table \ref{tab:PerformanceEva} compares our results with other published symbols spotting approaches. In this context, the evaluation metrics in the literature differ from the ones commonly used for assessing object detection networks, and are computed instance-wise and pixel-wise. For the instance-wise metrics, detected symbols that have some overlap with the ground truth are all counted as positive detections, and precision, recall and F-score values are calculated accordingly. Pixel-wise metrics, based on relevant and non-relevant retrieved pixels, refine the localization assessment \cite{rusinol2009performance}. In Table \ref{tab:PerformanceEva}, $P$, $R$ and $F$ stand for precision, recall and F-score, respectively. The `Queries' column indicates how many of the 16 symbols in the dataset are employed in the evaluation. The winning method in \cite{valveny2011report} was \cite{nayef2011use}. Our method significantly outperforms all other methods; the one method with comparable performance ($5^{th}$ row) was evaluated on a limited subset of the symbols only.

\begin{figure}[t!]
\centering
\includegraphics[width = 0.42\textwidth, frame]{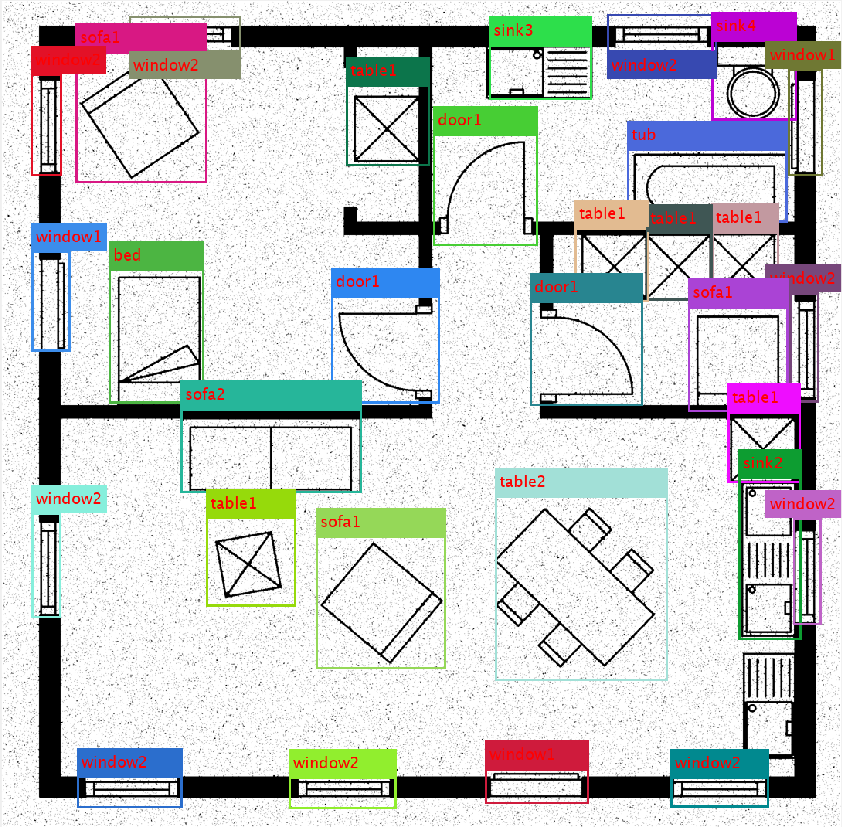}
\caption{Examples of spotted symbols on a degraded SESYD floor plan (noise 3).}
\label{Fig:SESYDPlans}
\end{figure}

Fig.\@ \ref{Fig:SESYDPlans} shows examples of spotted symbols on a SESYD layout for the noise \#3 degraded version. All symbol instances are correctly detected except for one (bottom right). Although SESYD does not include graphical notation variability, occlusion nor clutter as the real-world dataset does, it does allow us to conclude that our approach is able to deal with various degradation and noise levels.

\section{Conclusion}
\label{Sec:Conclusion}

This paper proposes a novel approach to symbol spotting utilizing a deep learning-based framework, showcased on the challenging application of real-world digital architectural floor plan analysis. We adapt an object detection network based on the YOLO architecture, and propose a training strategy based on tiles, allowing us to address many issues of the network regarding the relative small size of symbols compared to entire floor plans, aspect ratios, and data augmentation. Experiments on a dataset of real-world floor plans demonstrate that our proposed method successfully spots symbols in conditions under which traditional symbol spotting methods cannot cope, i.e.\@ symbols with low intra-class similarity and of variable graphical complexity, even in the presence of occlusion and clutter. The ResNet50 backbone within the YOLO framework yields the best results compared to the original Darknet19 and Xception backbones. Additional experiments on the public SESYD dataset also confirm that our method can deal with various degradation and noise levels and outperforms existing symbol spotting methods. Future research directions include the integration of contextual information relating to walls and rooms to further improve the detection results. We are also currently in the process of securing permissions from various architectural firms to release a public dataset of real-world architectural plans.

\section{Acknowledgement}
This research was supported by NSERC Canada and Triumph Electrical Consulting Engineering Ltd.\@ through the CRD Grants Program. The authors thank Steven Cooke at Triumph for providing the real-world dataset and for his help in interpreting architectural drawings.

{\small
\bibliographystyle{unsrt}
\bibliography{egbib}
}

\end{document}